\documentclass[letterpaper]{article} % DO NOT CHANGE THIS
\usepackage{aaai2026}  % DO NOT CHANGE THIS
\usepackage{times}  % DO NOT CHANGE THIS
\usepackage{helvet}  % DO NOT CHANGE THIS
\usepackage{courier}  % DO NOT CHANGE THIS
\usepackage[hyphens]{url}  % DO NOT CHANGE THIS
\usepackage{graphicx} % DO NOT CHANGE THIS
\usepackage{amsmath}
\usepackage{natbib}  % DO NOT CHANGE THIS AND DO NOT ADD ANY OPTIONS TO IT
\usepackage{caption} % DO NOT CHANGE THIS AND DO NOT ADD ANY OPTIONS TO IT
\usepackage{multirow} 
\usepackage{xcolor}
\usepackage{array}
\usepackage{algorithm}
\usepackage{algorithmic}

\usepackage{newfloat}
\usepackage{listings}
\DeclareCaptionStyle{ruled}{labelfont=normalfont,labelsep=colon,strut=off} % DO NOT CHANGE THIS
\floatstyle{ruled}
\newfloat{listing}{tb}{lst}{}
\floatname{listing}{Listing}
\title{MA-CBP: A Criminal Behavior Prediction Framework Based on\\Multi-Agent Asynchronous Collaboration}
\author{
    Cheng Liu\equalcontrib\textsuperscript{\rm 1},
    Daou Zhang\equalcontrib\textsuperscript{\rm 2},
    Tingxu Liu\equalcontrib\textsuperscript{\rm 1},
    Yuhan Wang\equalcontrib\textsuperscript{\rm 1},
    Jinyang Chen\textsuperscript{\rm 3},
    Yuexuan Li\textsuperscript{\rm 2},
    Xinying Xiao\textsuperscript{\rm 1},
    Chenbo Xin\textsuperscript{\rm 1},
    Ziru Wang\textsuperscript{\rm 4},
    Weichao Wu\textsuperscript{\rm 1}\thanks{Corresponding author.},
}
\affiliations{
    \textsuperscript{\rm 1}School of Mechatronical Engineering, Beijing Institute of Technology,\\
    \textsuperscript{\rm 2}School of Automation, Beijing Institute of Technology,\\
    \textsuperscript{\rm 3}School of Computer Science and Technology, Beijing Institute of Technology,\\
    \textsuperscript{\rm 4}School of Information and Electronics, Beijing Institute of Technology,\\
    \{3120235471, 1120211575, 3120240175, 3120240276, wuweichao\}@bit.edu.cn
}

\usepackage{bibentry}
\begin{document}

\maketitle

\begin{abstract}
With the acceleration of urbanization, criminal behavior in public scenes poses an increasingly serious threat to social security. Traditional anomaly detection methods based on feature recognition struggle to capture high-level behavioral semantics from historical information, while generative approaches based on Large Language Models (LLMs) often fail to meet real-time requirements. To address these challenges, we propose MA-CBP, a criminal behavior prediction framework based on multi-agent asynchronous collaboration. This framework transforms real-time video streams into frame-level semantic descriptions, constructs causally consistent historical summaries, and fuses adjacent image frames to perform joint reasoning over long- and short-term contexts. The resulting behavioral decisions include key elements such as event subjects, locations, and causes, enabling early warning of potential criminal activity. In addition, we construct a high-quality criminal behavior dataset that provides multi-scale language supervision, including frame-level, summary-level, and event-level semantic annotations. Experimental results demonstrate that our method achieves superior performance on multiple datasets and offers a promising solution for risk warning in urban public safety scenarios.
\end{abstract}

\begin{links}
    \link{Code}{https://ma-cbp.github.io/}
    % \link{Extended version}{https://arxiv.org/abs/}
\end{links}

\section{Introduction}

As urbanization accelerates and public spaces such as streets, shops, and transportation hubs become increasingly open, security risks in urban environments are rising. Criminal activities such as theft, robbery, arson, and explosions pose serious threats to public order and the safety of lives and property. Leveraging artificial intelligence to efficiently and accurately identify criminal intent has become a core challenge for intelligent surveillance and security systems~\cite{chalapathy2019deep}.

Criminal behavior prediction fundamentally involves analyzing actions that deviate significantly from normal patterns in a given context. Since criminal acts are often preceded by behavioral cues that differ from those in normal environments, predictive methods can leverage feature detection and deep reasoning to identify high-risk abnormal behaviors~\cite{sultani2018real}. Therefore, criminal behavior prediction not only focuses on the anomalies themselves but also emphasizes modeling and reasoning about underlying motivations, behavioral chains, and precursor events—representing a high-dimensional extension of traditional anomaly detection.

\begin{figure}[htb]
\centering
\includegraphics[width=1\linewidth]{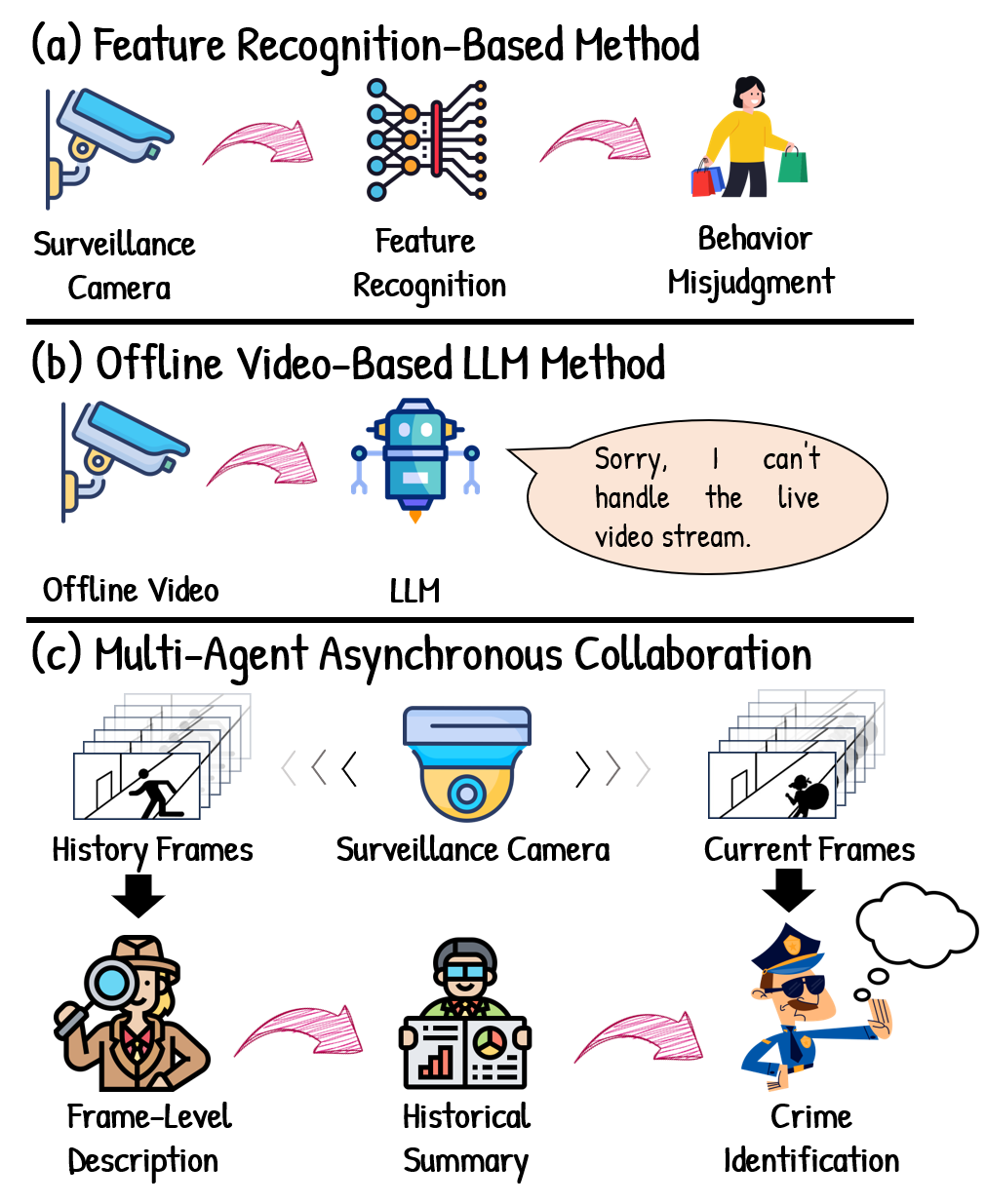}
\caption{Multi-Agent Asynchronous Collaboration (Ours) vs. existing methods. (a) Feature recognition-based methods detect abnormal behavior by analyzing behavior features. (b) Offline Video-Based LLM Method cannot process real-time video streams. (c) Our method enables early warning of criminal behavior by translating real-time video streams into textual semantics.}
\label{fig1}
\end{figure}

From the perspective of dynamic behavioral evolution, abnormal behaviors can be categorized into transient anomalies and persistent anomalies, depending on whether a perceptible evolution process precedes their occurrence. Transient anomalies are typically characterized by abrupt, drastic, and discontinuous behavioral changes, often triggered by sudden external events, such as explosions or acts of arson~\cite{yang2025assistpda,jin2023video}. Although this type of anomaly is well-suited for traditional real-time detection models, it remains challenging to predict and intervene in advance. In contrast, persistent anomalies often exhibit observable precursory signals, with their behavioral evolution patterns gradually emerging over time through time series, behavioral trajectories or semantic contexts. For example, shoplifting is often accompanied by a series of indicative dynamic features before it occurs, such as frequent wandering between shelves during non-peak hours, frequent observation of surveillance cameras, abnormal hand movements to cover up, and repeated viewing of a certain product~\cite{s21206812,martinez2021criminal}. Although these detailed features may belong to the category of normal behavior when viewed individually, they show significant deviations from the normal state in the time series under a specific background. Therefore, early warning of criminal behavior can be achieved by identifying the potential correlation between these preceding signals and the final behavior.

Traditional architectures based on Convolutional Neural Network (CNN) or Recurrent Neural Network (RNN) often struggle to achieve ideal prediction results when dealing predictable criminal behaviors. This is mainly because their ability to semantically model the process of behavioral changes is limited. For instance, non-semantic actions are first detected using 3D-CNN, and then action sequence analysis is performed on these detected actions to convert non-semantic actions into semantic information, thereby enabling the recognition of theft behaviors and prediction of criminal intent~\cite{kim2021identifying}. However, deep learning methods based on low-level feature recognition are unable to capture high-level behavioral semantics and generally have difficulty adapting to new environments or unseen behavior patterns, as shown in Figure~\ref{fig1}(a). In contrast, Large Language Models (LLMs) can effectively mine potential prior signals embedded in the evolution of abnormal behaviors, thus improving the prediction accuracy of persistent anomalies. For example, Holmes-VAD and Holmes-VAU~\cite{zhang2024holmes,zhang2025holmes} integrate visual features with language information and leverage LLMs to perform semantic reasoning on behaviors, achieving both high anomaly detection accuracy and strong explainability. However, LLM-based Video Anomaly Detection (VAD) methods mainly depend on offline video input for global analysis and must wait for the entire video segment to be loaded before reasoning and making judgments. This limitation prevents their deployment in security monitoring systems that require high real-time responsiveness, as shown in Figure~\ref{fig1}(b). In summary, current approaches exhibit clear shortcomings in the task of criminal behavior detection based on abnormal behavior. Models that can respond in real time often lack a deep understanding of behavioral context, while those capable of constructing rich semantic contexts struggle to meet the real-time processing requirements necessary for online applications.                                                     

To reconcile the contradiction between real-time responsiveness and context understanding, this paper transforms visual modalities into language representations to enable early warning of criminal behavior, as shown in Figure~\ref{fig1}(c). Specifically, we propose a criminal behavior prediction framework based on multi-agent asynchronous collaboration. The framework first generates frame-level semantic description of video streams, then constructs a causally consistent historical summary, and finally performs joint reasoning over both short-term and long-term temporal windows by fusing adjacent image frames. Furthermore, we build a multi-scale, high-quality criminal behavior dataset to enhance the accuracy and explainability of criminal behavior prediction.                                                                                                         

Our main contributions can be summarized as follows:

\begin{itemize}
    \item We propose a pioneering early warning framework for criminal behavior based on LLMs, overcoming the limitations of existing methods that fail to simultaneously achieve deep understanding of historical behavior semantics and meet the requirements for practical deployment.
    \item We propose a real-time reasoning agent, which performs joint short-term and long-term inference by integrating historical summaries with current visual information, and generates structured decisions regarding potential criminal behavior.
    \item We have constructed a high-quality criminal behavior dataset that covers four categories of abnormal events and includes fine-grained natural language annotations, providing strong data support for research on criminal behavior prediction.
\end{itemize}

% You can remove the copyright notice and ensure that your names aren't shown by including \texttt{submission} option when loading the \texttt{aaai2026} package:

% \begin{quote}\begin{scriptsize}\begin{verbatim}
% \documentclass[letterpaper]{article}
% \usepackage[submission]{aaai2026}
% \end{verbatim}\end{scriptsize}\end{quote}

\section{Related Work}
\subsubsection{Criminal Behavior Dataset.}

In recent years, a variety of benchmark datasets for video analysis have emerged in the field of criminal behavior detection, covering diverse application environments from posture information to multimodal perception, and from retail scenarios to urban public spaces. For example, the PoseLift dataset~\cite{rashvand2025exploring} is a privacy-preserving dataset designed for theft detection in retail security, derived from anonymized human pose data collected in real retail environments. The CamNuvem dataset~\cite{de2022camnuvem} focuses on robbery behavior and contains videos recorded in various real-world scenarios. The UCF-Crime dataset~\cite{sultani2018real} is a large-scale, real-world abnormal event detection benchmark that covers 13 common urban public safety incidents, such as shoplifting, fighting, robbery, arson, and more. It is widely used in anomaly detection tasks, particularly  for evaluating the model's ability to identify and differentiate high-risk behaviors in unstructured environments. The XD-Violence dataset~\cite{wu2020not} contains over 2,000 videos of violent behaviors captured in real or semi-real scenarios, covering various abnormal events such as fighting and robbery. This dataset integrates visual, audio, and motion modalities, making it well-suited for multimodal violent behavior recognition tasks. The criminal behavior dataset used in this paper is derived from the UCF-Crime and CamNuvem datasets. The data is divided into two stages: suspicious behavior segments preceding the crime and typical behavior segments during the commission of the crime. This forms a two-stage sample set that includes both abnormal behavior and criminal behavior.

\subsubsection{Criminal Behavior Prediction.}

In recent years, existing studies have explored the prediction of criminal behaviors, in surveillance videos from multiple perspectives, including human posture, visual spatiotemporal features, and motion trajectories. For example, based on the PoseLift dataset, a transformer-based architecture called Shopformer~\cite{rashvand2025shopformer} reduces reliance on raw video by processing pose data. However, modeling theft solely as a human posture detection task may lead to misjudgment of certain atypical but legal behaviors. Other studies have adopted deep learning methods based on spatiotemporal feature fusion, extracting spatial features using CNNs and modeling temporal dynamics with RNNs to identify theft in videos videos~\cite{muneer2023shoplifting,ansari2022esar,ansari2022expert}. However, these methods lack interpretable analysis of the model's decision-making process, making the rationale behind their predictions difficult to understand and unsuitable for deployment in security-sensitive environments. To reduce reliance on high computational resources, some recent studies have instead used only the bounding box coordinates of individuals in the video to construct temporal feature sequences~\cite{nazir2023suspicious}. While this approach is more lightweight, it fails to capture fine-grained movements and results in low prediction accuracy for complex criminal behaviors. Different from the above-mentioned feature recognition-based methods, this paper constructs frame-level semantic descriptions and historical summaries from real-time video streams to form a long-term memory of abnormal behavior patterns. It then combines the short-term states of adjacent current frames to enable timely responses to potential criminal behavior.

% \begin{itemize}
% \item You must use the 2026 AAAI Press \LaTeX{} style file and the aaai2026.bst bibliography style files, which are located in the 2026 AAAI Author Kit (aaai2026.sty, aaai2026.bst).
% \item You must complete, sign, and return by the deadline the AAAI copyright form (unless directed by AAAI Press to use the AAAI Distribution License instead).
% \item You must read and format your paper source and PDF according to the formatting instructions for authors.
% \item You must submit your electronic files and abstract using our electronic submission form \textbf{on time.}
% \item You must pay any required page or formatting charges to AAAI Press so that they are received by the deadline.
% \item You must check your paper before submitting it, ensuring that it compiles without error, and complies with the guidelines found in the AAAI Author Kit.
% \end{itemize}

\section{Method}

\begin{figure}[htb] 
 \centering 
 \includegraphics[width=1\linewidth]{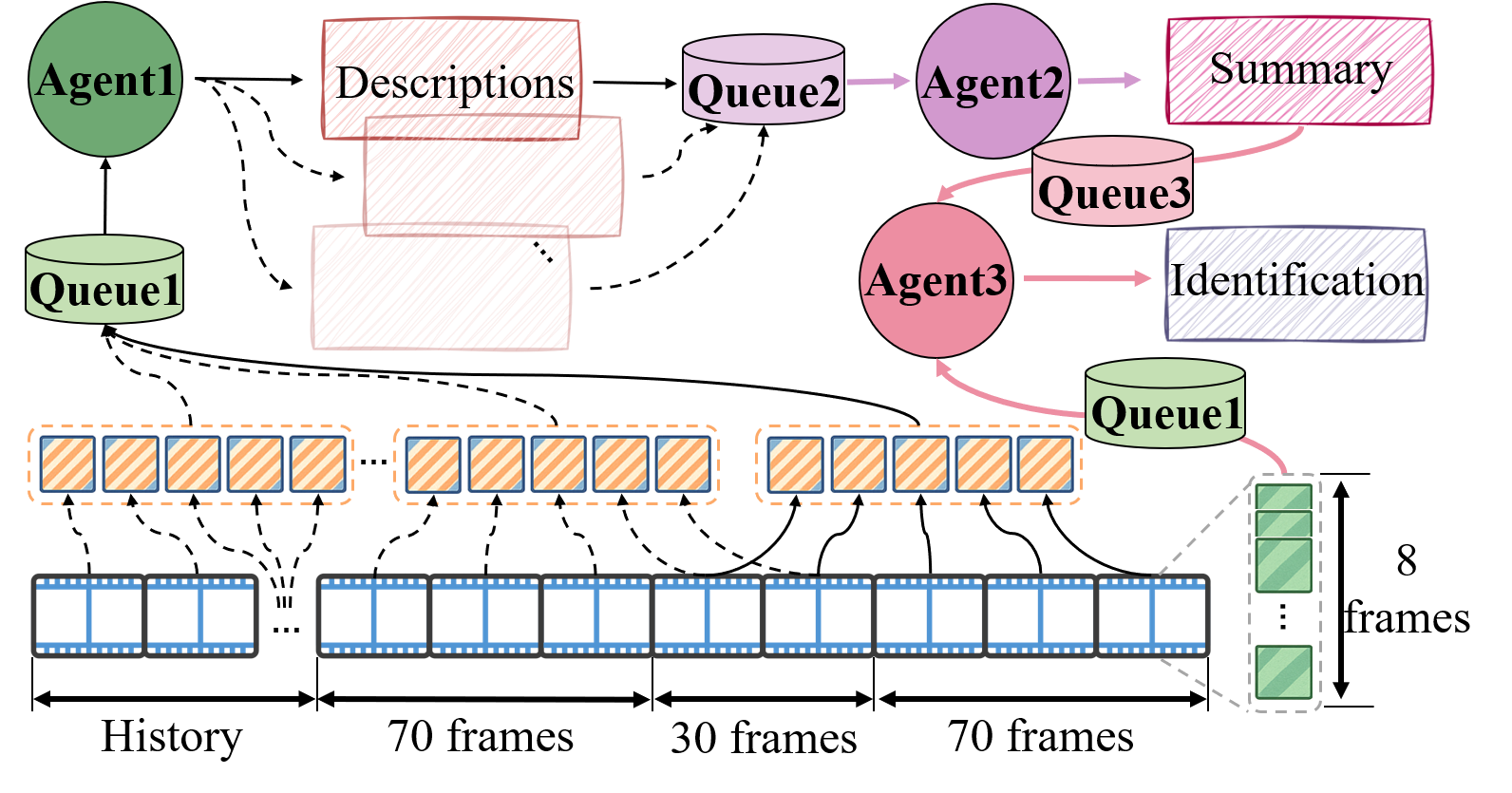} 
 \caption{MA-CBP workflow. For every 100 video frames, Agent 1 uniformly samples 5 frames to generate semantic descriptions (sampling occurs every 70 frames to ensure overlap between adjacent 100-frame segments). Agent 2 combines these 5 descriptions with all previous summaries to update the latest summary, which accumulates the full historical context. Agent 3 continuously receives the most recent summary along with 8 image frames, sampled every 0.1 seconds from the current timestamp.} 
 \label{fig2} 
\end{figure}

We propose MA-CBP, a criminal behavior prediction framework based on multi-agent asynchronous collaboration. As shown in Figure~\ref{fig2}, MA-CBP comprises three agents and three ZeroMQ-based message queues.

% \subsection{Model Details}
\subsubsection{Language Representation Based on Visual Information.}

Humans often understand images by converting visual information into language representations, thereby constructing cognitive models with semantic structures~\cite{marr2010vision,bisk2020experience}. On the one hand, this cognitive mechanism inspires the language-mediated understanding pathways in computational models, enabling the system to characterize key visual events in time series using a language modality, thereby providing a foundation for the construction of semantic continuity and causal chains for intentional reasoning. On the other hand, by translating images into interpretable language representations, the computational burden of processing high-dimensional visual data can be significantly reduced.

This paper selects BLIP~\cite{li2022blip} as the core architecture of the frame-level semantic description model to generate frame-level semantics for image frames in video stream. BLIP, which is based on the BERT architecture, features fast inference speed and is well suited for tasks requiring high real-time performance. Specifically, we extract 5 image frames from the latest 100 frames and generate frame-level semantic descriptions. For each frame $I_i$, we compute $g_i, t_i = \mathcal{M}_{BLIP}(I_i)$, where $\mathcal{M}_{BLIP}(\cdot)$ is the model mapping function, $g_i$ denotes the semantic representation, and $t_i$ is the corresponding timestamp. 

\subsubsection{Redundant Information Filtering Mechanism.}

In the language representation task based on visual information, single-frame semantic representations are prone to semantic ambiguity and redundancy. To improve the consistency of semantic descriptions across consecutive frames, this paper proposes a main entity screening and redundant description filtering mechanism based on entity frequency distribution.

Let the predefined entity vocabulary be $V=\{e_1, e_2, \ldots, e_n\}$, where the occurrence frequency of each entity word $e_i$ in the vocabulary is denoted as $N_i$. We define the frequency set as $N=\{N_1, N_2, \ldots, N_k\}$ and sort it in descending order such that $N_1 \geq N_2 \geq ... \geq N_k$. First, define the semantic main entity set is defined as $S \subseteq V $, and a hierarchical screening criterion is constructed as $\mathcal{S} = \{e_i \mid N_i > \tau \land N_{i-1} > 2 \cdot N_i\}$, where $\tau = 3$ is the minimum frequency threshold for retention. Second, the redundant entity set is defined as a difference set $\mathcal{D} = S \setminus V $, and an information redundancy filtering function $f(g)$ is introduced. If a frame-level description $g$ contains only entities from $\mathcal{D}$, it is identified as semantically redundant and removed. Through this screening and filtering process, semantic ambiguity is effectively reduced, and key entities with strong semantic relevance are preserved, thereby improving the coherence of semantic expressions across image frames.

\subsubsection{Historical Summary Model.}

Although the frame-level description generated by BLIP has good semantic explainability, they are insufficient to directly facilitate modeling of complex dynamic behaviors. To enhance the semantic aggregation ability of consecutive frames information and improve behavior understanding, this paper uses prompt engineering to automatically generate summary-level descriptions of video clips. Specifically, let the image descriptions be denoted as $\mathcal{G} = \{g_i\}_{i=1}^{N_g}$, where $N_g$ represents the length of the cache sequence of single-frame descriptions, empirically set to 5. Given a designed prompt, the final historical summary $h_s$ is computed as:

\begin{equation}
h_s = \mathcal{M}_{API}(prompt, f(\mathcal{G}), h_l),
\end{equation}

\noindent where $\mathcal{M}_{API}$ denotes the Qwen API~\cite{wan2025qwenlong}, and $h_l$ is the previous summary.

\subsubsection{Criminal Behavior Discrimination Model.}

\begin{figure}[htb]
\centering
\includegraphics[width=0.95\columnwidth]{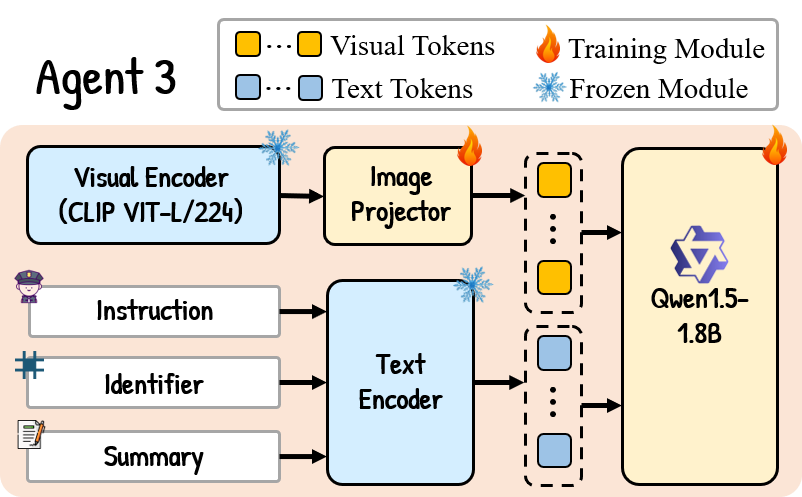}
\caption{Model architecture of Agent 3. When Agent 2 outputs summary, visual encoder immediately receives 8 image frames, while the text encoder processes the instruction, identifier, and summary. The identifier explicitly marks the semantic boundaries between consecutive frames using the frame separator \texttt{<sep>}, and encapsulates the historical summary using \texttt{<history>}. After obtaining the visual and language embeddings output from the visual and text encoders, the embeddings are concatenated and passed to Qwen1.5-1.8B to generate the discrimination result.}
\label{fig3}
\end{figure}

In response to the challenges of limited context windows, insufficient temporal modeling capabilities, and blurred module boundaries in traditional multimodal models when processing long temporal image sequences~\cite{liu2023llava}, this paper extends the model’s input paradigm from single-frame to batch-frame processing and explicitly defines semantic boundaries using delimiters. Furthermore, the language decoder is replaced with the Qwen1.5-1.8B~\cite{bai2023qwentechnicalreport} with a larger context window and faster processing speed, as shown in the Figure~\ref{fig3}.

For visual modality, let the sequence of adjacent image frames be denoted as $\mathcal{I}=\{I_1, I_2, \ldots, I_8\}$. This sequence is extracted with high-level semantic features through the frozen visual encoder $f_{vis}(\cdot)$ to obtain the image embedding $\mathcal{V} = \{v_1, v_2, \ldots, v_8\} = f_{vis}(\mathcal{I})$. Subsequently, the image features are modally aligned and dimensionally compressed by the image projector $P_\theta(\cdot)$ to obtain the cross-modal embedding $\mathcal{V}' = P_\theta(\mathcal{V})$, where $\theta$ denotes the trainable parameters of the projector.

For the language modality, the text input consists of task instructions $\mathcal{T}_{inst}$, semantic identifiers $\mathcal{T}_{id}$, and historical summary content $\mathcal{T}_{sum}$, which are concatenated to the complete text $\mathcal{T} = \mathcal{T}_{inst} \oplus \mathcal{T}_{id} \oplus \mathcal{T}_{sum}$. The combined text information is mapped to a language embedding $\mathcal{L} = f_{text}(\mathcal{T})$ through a frozen text encoder $f_{text}(\cdot)$ for subsequent language reasoning.

Finally, the visual modality embedding $\mathcal{V}$ and the language embedding $\mathcal{L}$ are combined into a unified representation $\mathcal{U} = \mathcal{L} \oplus \mathcal{V}'$, which is provided as a prompt to the Qwen1.5-1.8B $\mathcal{M}_{Qwen}$ to perform the conditional language generation:

\begin{equation}
\hat{y} = \mathcal{M}_{Qwen}(\mathcal{U}) = \mathcal{M}_{Qwen}(f_{text}(\mathcal{T}) \oplus P_\theta(f_{vis}(\mathcal{I}))),
\end{equation}

\noindent where $\hat{y}$ is a structured discrimination result, including the behavior subject, the location and the potential motivation.

% 修正表格1的位置和格式
\begin{table*}[!ht]
\centering
\setlength{\tabcolsep}{2.5mm}
\begin{tabular}{l|c|cccc|cccc|cccc}
\hline
\multirow{2}{*}{\textbf{Method}} & \multirow{2}{*}{\textbf{Params}} & \multicolumn{4}{c|}{\textbf{F1-score ↑}} & \multicolumn{4}{c|}{\textbf{AUC ↑}} & \multicolumn{4}{c}{\textbf{AP ↑}} \\
\cline{3-14}
 & & \textbf{St} & \textbf{Ro} & \textbf{Bu} & \textbf{Sh} & \textbf{St} & \textbf{Ro} & \textbf{Bu} & \textbf{Sh} & \textbf{St} & \textbf{Ro} & \textbf{Bu} & \textbf{Sh} \\
\hline
VideoLLaMA2 & 7B & 0.57 & 0.58 & 0.56 & 0.49 & 0.51 & 0.51 & 0.52 & 0.49 & 0.45 & 0.49 & 0.44 & 0.40 \\
InternVL2.5 & 2B & 0.00 & 0.15 & 0.49 & 0.09 & 0.50 & 0.54 & 0.43 & 0.52 & 0.44 & 0.52 & \textbf{0.98} & 0.43 \\
Qwen2.5-VL & 3B & \textbf{0.77} & 0.77 & \textbf{0.64} & \textbf{0.73} & 0.78 & 0.75 & 0.67 & \textbf{0.77} & 0.66 & 0.67 & 0.54 & 0.62 \\
LLaVA-NeXT & 7B & 0.05 & 0.39 & 0.11 & 0.08 & 0.50 & 0.61 & 0.50 & 0.49 & 0.44 & 0.58 & 0.43 & 0.40 \\
Video-LLaVA & 7B & 0.46 & 0.70 & 0.64 & 0.60 & 0.52 & 0.71 & 0.67 & 0.65 & 0.46 & 0.64 & 0.54 & 0.50 \\
Holmes-VAD & 7B & 0.10 & 0.28 & 0.30 & 0.00 & 0.51 & 0.57 & 0.57 & 0.50 & 0.45 & 0.55 & 0.49 & 0.41 \\
Holmes-VAU & 2B & 0.25 & 0.26 & 0.40 & 0.14 & 0.57 & 0.57 & 0.63 & 0.57 & 0.52 & 0.55 & 0.57 & 0.49 \\
LAVAD & 13B & 0.11 & 0.06 & 0.11 & 0.00 & 0.53 & 0.50 & 0.52 & 0.50 & 0.47 & 0.48 & 0.45 & 0.41 \\
\hline
MA-CBP & 1.8B & 0.72 & \textbf{0.87} & 0.63 & 0.63 & \textbf{0.79} & \textbf{0.88} & \textbf{0.72} & 0.73 & \textbf{0.76} & \textbf{0.87} & 0.67 & \textbf{0.68} \\
\hline
\end{tabular}
\caption{Detection performance comparison with the state-of-the-art methods. "St" stands for stealing, "Ro" stands for robbery, "Bu" stands for burglary, "Sh" stands for shoplifting.}
\label{tab1}
\end{table*}

\section{Dataset Benchmark}
\subsubsection{Data Collection.}
Firstly, the original videos were manually screened to remove samples with poor image quality or unclear semantics. Secondly, four categories of abnormal videos, namely shoplifting, stealing, robbery, and burglary, were selected from the UCF-Crime and CamNuvem datasets, resulting in a total of 711 videos. At the same time, 950 high-quality normal videos were selected from MSVD~\cite{chen2011collecting}. Finally, abnormal video segments were extracted using our self-developed annotation tool, which identifies clips based on suspicious behaviors preceding the crime and typical actions during the criminal event.

\subsubsection{Dataset Construction Process.}
In the frame-level description generation stage, we employ GPT-4.1~\cite{achiam2023gpt} to guide BLIP in generating semantic descriptions for each frame, ensuring that the descriptions align more closely with human cognition. In the historical summary generation stage, to guide the model to build contextual causal relationships, we integrate frame-level semantic information and previous summaries to generate coherent summary data. In the event segment description generation stage, GPT-4.1 is used to automatically generate text descriptions for each event segment, focusing on extracting the location of the event, the subject of the behavior, and the rationale behind the judgment. All generated results were manually reviewed, resulting in a high-quality semantic description set.  

\section{Experiments}
\subsection{Experimental Setup}
\subsubsection{Datasets.}

In this experiment, LLaVA-CC3M-Pretrain-595K~\cite{liu2023llava} is used for vision-language alignment pre-training, while MSVD is utilized for multi-frame semantic modeling fine-tuning. High-quality samples selected from UCF-Crime, CamNuvem, and MSVD are then used for history-enhanced reasoning fine-tuning, and are divided into 1,130 training samples and 531 test samples.

\subsubsection{Evaluation Metrics.}
In the comparative experiments, we selected four typical scenes as evaluation objects and used F1-score, Area Under ROC Curve (AUC) and Average Precision (AP) to quantitatively assess the model's performance based on the correspondence between each video's classification result and its ground-truth label. In the ablation experiments, we employed a variety of mainstream text generation evaluation indicators, including BLEU, CIDEr, ROUGE, METEOR and Latency, to comprehensively evaluate the model's generation quality and speed.  

\subsubsection{Implementation Details}

To enhance visual-language alignment, temporal modeling, and cross-modal reasoning for criminal behavior prediction, we propose a three-stage training framework. 

The first stage is the visual-language alignment pre-training, which aims to guide Agent 3 in building a unified embedding space between the visual and language modalities. This stage is trained for 5 epochs on the LLaVA-CC3M-Pretrain-595K dataset, with only the multimodal projection layer parameters kept trainable.

The second stage is multi-frame semantic modeling fine-tuning, which aims to enhance Agent 3's ability to understand the contextual relationship and causal logic within image sequence. In this stage, 8 image frames are uniformly sampled from full videos in MSVD to construct feature representations of consecutive frames. The Low-Rank Adaptation (LoRA)~\cite{hu2021loralowrankadaptationlarge} method is employed to fine-tune the projection layer and Qwen1.5-1.8B over 5 epochs.

The third stage is history-enhanced reasoning fine-tuning, which aims to incorporate historical behavior information to improve the model's reasoning ability for predictable criminal behaviors. In this stage, abnormal behavior segments are selected from the UCF-Crime and CamNuvem datasets, while normal segments drawn from the MSVD dataset. From each segment, 100 frames are uniformly sampled to construct temporal information. To enhance the temporal generalization ability of the model and augment the sample size, these 100 frames are divided into historical and adjacent segments according to different ratios (3:7, 1:1 and 7:3). Finally, 8 image frames are uniformly sampled again from the adjacent segments. The training process employs the LoRA method to fine-tune the projection layer and Qwen1.5-1.8B over 6 epochs.

\subsection{Main Results}
\subsubsection{Results on Video Reasoning.}

Existing research on crime behavior prediction mostly relies on private data or closed evaluation systems, lacking reproducibility support, which makes fair comparison~\cite{hasan2016learning,wang2019godsgeneralizedoneclassdiscriminative,thakare2023dyannet,wu2024vadclip}. Since the training and testing details of other models are unavailable, we evaluate all comparison models on a mixed dataset comprising UCF-Crime, CamNuvem, and MSVD to ensure fairness. To comprehensively evaluate the actual performance of our model, we compare the proposed method with the current state-of-the-art generative methods, including explainable multi-modal methods~\cite{cheng2024videollama2advancingspatialtemporal,chen2025expandingperformanceboundariesopensource,Qwen2.5-VL,zhang2024llavanext-video,lin2024videollavalearningunitedvisual}, weakly-supervised methods~\cite{zhang2024holmes,zhang2025holmes}, and training-free methods~\cite{zanella2024harnessing}. Table~\ref{tab1} summarizes the detection performance of each model on four typical predictable criminal behaviors. Experimental results show that compared with other generative methods, MA-CBP achieves superior performance on criminal behavior detection despite having only 1.8B parameters, which is significantly smaller than most multimodal models with 7B parameters or more. Notably, in the robbery category, our method significantly outperforms other generative models, achieving an F1-score of 0.87, AUC of 0.88, and AP of 0.87.

\subsubsection{Results on Early-Stage Video Reasoning.}

We designed an early warning experiment that provides the model with only the segments between the onset of abnormal behavior and the clear occurrence of criminal behavior. For fairness, we compare our method only with VAD models. The experimental results are presented in Table~\ref{tab2}. Despite using only abnormal behavior segments, our method effectively captures potential risk behaviors and demonstrates superior capability in early prediction of criminal behavior.

\begin{table}[htp]
\centering
\setlength{\tabcolsep}{4mm}
\begin{tabular}{l|ccc}
\hline
\textbf{Method} & \textbf{F1-score ↑} & \textbf{AUC ↑} & \textbf{AP ↑} \\
\hline
Holmes-VAD & 0.19 & 0.54 & 0.50 \\
Holmes-VAU & 0.36 & 0.61 & 0.58 \\
LAVAD & 0.08 & 0.51 & 0.47 \\
MA-CBP & \textbf{0.88} & \textbf{0.89} & \textbf{0.88} \\
\hline
\end{tabular}
\caption{Evaluation of reasoning performance in early-stage video segments prior to crime onset.}
\label{tab2}
\end{table}

\subsubsection{Qualitative Comparision.}

\begin{figure}[!h] 
 \centering 
 \includegraphics[width=1\linewidth]{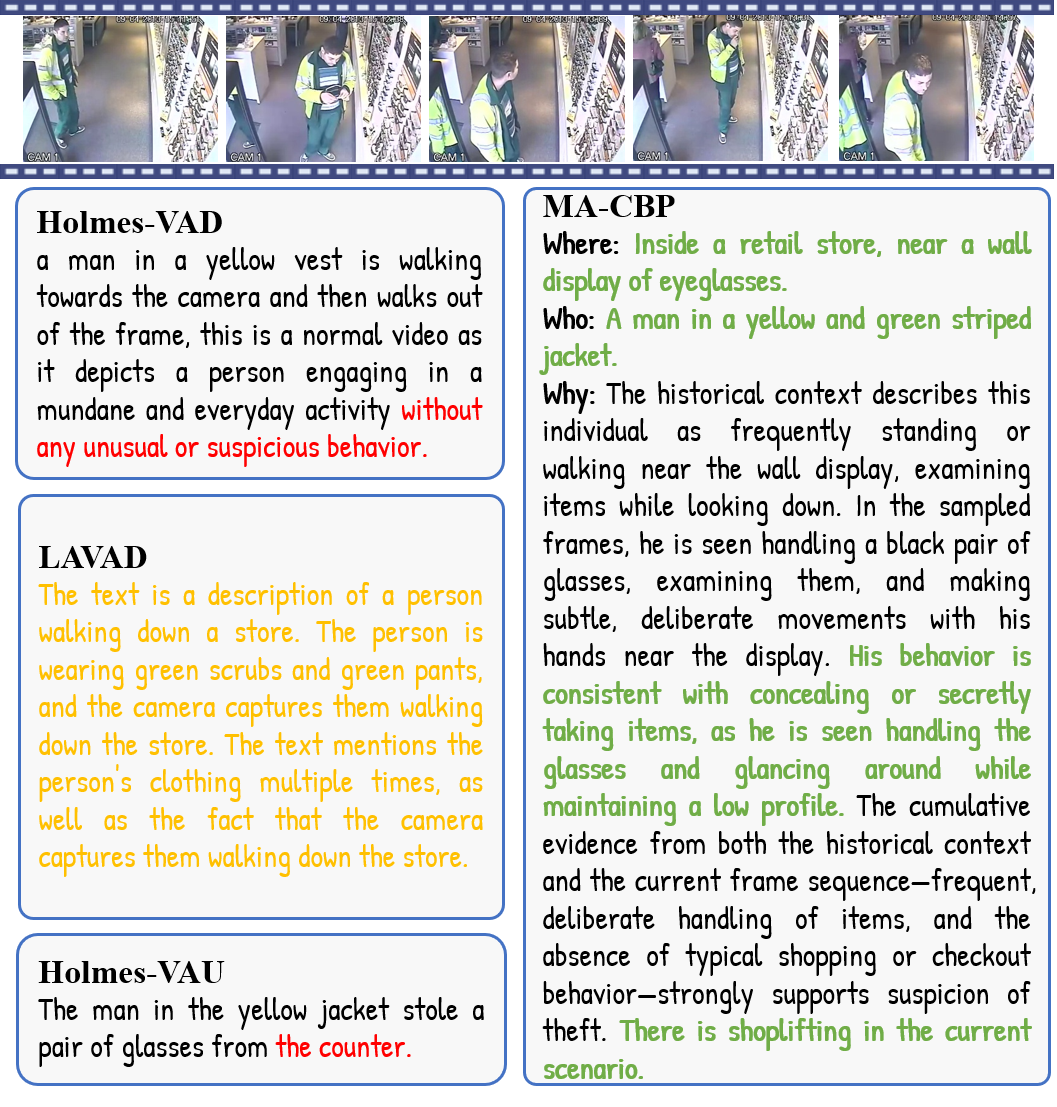} 
 \caption{Qualitative comparison of different anomaly detection methods on criminal behavior detection. In the model's answers, correct, unimportant, and incorrect explanations are highlighted in green, yellow, and red, respectively.} 
 \label{fig4} 
\end{figure}

We provide a qualitative comparison of our model with other anomaly detection models, as shown in the Figure~\ref{fig4}. The results demonstrate that our model can accurately predict possible criminal incidents in the video based on abnormal behaviors and provide corresponding explainable analysis and localization.

\subsection{Analysis}
\subsubsection{Effect of Sub-Agent.}

In Table~\ref{tab3}, we conduct an ablation study by removing Agent 2 and directly feeding the output of Agent 1 into Agent 3 (Note that since the output of Agent 1 is a necessary prerequisite for the input of Agent 2, Agent 1 cannot be ablated.). We refer to this variant as F-Only. Experimental results show that, due to the absence of contextual semantic integration, F-Only performs worse than MA-CBP across all evaluation metrics.

\begin{table}[htp]
\centering
\setlength{\tabcolsep}{1mm}
\begin{tabular}{l|cccc}
\hline
\textbf{Method} & \textbf{BLEU ↑} & \textbf{CIDEr ↑} & \textbf{ROUGE ↑} & \textbf{METEOR ↑} \\
\hline
F-Only & 1.152 & 0.405 & 0.474 & 0.423 \\
MA-CBP & \textbf{1.227} & \textbf{0.431} & \textbf{0.483} & \textbf{0.438} \\
\hline
\end{tabular}
\caption{Agent ablation performance. BLEU refers to the cumulative values from BLEU-1 to BLEU-4.}
\label{tab3}
\end{table}

\subsubsection{Effect of Real-Time Reasoning.}

We investigate the impact of different numbers of adjacent frames on model performance by varying the number of adjacent frames input to Agent 3. Specifically, we uniformly extract 4, 6, 8, and 10 frames from the adjacent segments and compare their effects on inference quality and speed. As shown in Table~\ref{tab4}, the model achieves better overall performance when using 8 adjacent frames. However, when the number increases to 10, the BLEU and Latency performances slightly declines. This may be due to the introduction of semantic redundancy or irrelevant information, which weakens the model’s ability to focus on key behavioral transitions. Based on this observation, we selected 8 frames in our final setup to ensure stronger contextual representation while minimizing information redundancy.

\begin{table}[htp]
\centering
\setlength{\tabcolsep}{2mm}
\begin{tabular}{c|ccc}
\hline
\textbf{Frames(N)} & \textbf{BLEU ↑} & \textbf{CIDEr ↑} & \textbf{Latency ↓ (s)} \\
\hline
4 & 1.220 & 0.417 & \textbf{3.3} \\
6 & 1.225 & 0.429 & 3.4 \\
8 & \textbf{1.227} & 0.431 & \textbf{3.3} \\
10 & 1.223 & \textbf{0.434} & 3.4 \\
\hline
\end{tabular}
\caption{Effect of different numbers of adjacent frames on performance. Latency refers to the minimum time required for our model to identify criminal behavior from the onset of abnormal behavior.}
\label{tab4}
\end{table}

\section{Conclusion}
In this paper, we propose a novel multi-agent asynchronous collaborative framework called MA-CBP for crime prediction in real-time video streams. By integrating frame-level semantics, historical summaries, and adjacent frames , our approach effectively bridges short-term visual with long-term behavioral semantics. Additionally, we introduce a high-quality dataset that supports multi-agent systems with interpretable data through multi-scale language supervision. Experiments on shoplifting, stealing, burglary, and robbery tasks demonstrate that our framework outperforms existing state-of-the-art generative and anomaly detection models. The results of our paper demonstrate that converting images into textual semantic representations is a highly effective and practically significant paradigm, contributing to the development of trustworthy intelligent surveillance systems.

% \bibliography{reference}

\begin{thebibliography}{36}
\providecommand{\natexlab}[1]{#1}

\bibitem[{Achiam et~al.(2023)Achiam, Adler, Agarwal, Ahmad, Akkaya, Aleman, Almeida, Altenschmidt, Altman, Anadkat et~al.}]{achiam2023gpt}
Achiam, J.; Adler, S.; Agarwal, S.; Ahmad, L.; Akkaya, I.; Aleman, F.~L.; Almeida, D.; Altenschmidt, J.; Altman, S.; Anadkat, S.; et~al. 2023.
\newblock Gpt-4 technical report.
\newblock \emph{arXiv preprint arXiv:2303.08774}.

\bibitem[{Ansari and Singh(2022{\natexlab{a}})}]{ansari2022esar}
Ansari, M.~A.; and Singh, D.~K. 2022{\natexlab{a}}.
\newblock ESAR, an expert shoplifting activity recognition system.
\newblock \emph{Cybern. Inf. Technol}, 22(1): 190--200.

\bibitem[{Ansari and Singh(2022{\natexlab{b}})}]{ansari2022expert}
Ansari, M.~A.; and Singh, D.~K. 2022{\natexlab{b}}.
\newblock An expert video surveillance system to identify and mitigate shoplifting in megastores.
\newblock \emph{Multimedia Tools and Applications}, 81(16): 22497--22525.

\bibitem[{Bai et~al.(2023)Bai, Bai, Chu, Cui, Dang, Deng, Fan, Ge, Han, Huang, Hui, Ji, Li, Lin, Lin, Liu, Liu, Lu, Lu, Ma, Men, Ren, Ren, Tan, Tan, Tu, Wang, Wang, Wang, Wu, Xu, Xu, Yang, Yang, Yang, Yang, Yao, Yu, Yuan, Yuan, Zhang, Zhang, Zhang, Zhang, Zhou, Zhou, Zhou, and Zhu}]{bai2023qwentechnicalreport}
Bai, J.; Bai, S.; Chu, Y.; Cui, Z.; Dang, K.; Deng, X.; Fan, Y.; Ge, W.; Han, Y.; Huang, F.; Hui, B.; Ji, L.; Li, M.; Lin, J.; Lin, R.; Liu, D.; Liu, G.; Lu, C.; Lu, K.; Ma, J.; Men, R.; Ren, X.; Ren, X.; Tan, C.; Tan, S.; Tu, J.; Wang, P.; Wang, S.; Wang, W.; Wu, S.; Xu, B.; Xu, J.; Yang, A.; Yang, H.; Yang, J.; Yang, S.; Yao, Y.; Yu, B.; Yuan, H.; Yuan, Z.; Zhang, J.; Zhang, X.; Zhang, Y.; Zhang, Z.; Zhou, C.; Zhou, J.; Zhou, X.; and Zhu, T. 2023.
\newblock Qwen Technical Report.
\newblock arXiv:2309.16609.

\bibitem[{Bai et~al.(2025)Bai, Chen, Liu, Wang, Ge, Song, Dang, Wang, Wang, Tang, Zhong, Zhu, Yang, Li, Wan, Wang, Ding, Fu, Xu, Ye, Zhang, Xie, Cheng, Zhang, Yang, Xu, and Lin}]{Qwen2.5-VL}
Bai, S.; Chen, K.; Liu, X.; Wang, J.; Ge, W.; Song, S.; Dang, K.; Wang, P.; Wang, S.; Tang, J.; Zhong, H.; Zhu, Y.; Yang, M.; Li, Z.; Wan, J.; Wang, P.; Ding, W.; Fu, Z.; Xu, Y.; Ye, J.; Zhang, X.; Xie, T.; Cheng, Z.; Zhang, H.; Yang, Z.; Xu, H.; and Lin, J. 2025.
\newblock Qwen2.5-VL Technical Report.
\newblock \emph{arXiv preprint arXiv:2502.13923}.

\bibitem[{Bisk et~al.(2020)Bisk, Holtzman, Thomason, Andreas, Bengio, Chai, Lapata, Lazaridou, May, Nisnevich et~al.}]{bisk2020experience}
Bisk, Y.; Holtzman, A.; Thomason, J.; Andreas, J.; Bengio, Y.; Chai, J.; Lapata, M.; Lazaridou, A.; May, J.; Nisnevich, A.; et~al. 2020.
\newblock Experience grounds language.
\newblock \emph{arXiv preprint arXiv:2004.10151}.

\bibitem[{Chalapathy and Chawla(2019)}]{chalapathy2019deep}
Chalapathy, R.; and Chawla, S. 2019.
\newblock Deep learning for anomaly detection: A survey.
\newblock \emph{arXiv preprint arXiv:1901.03407}.

\bibitem[{Chen and Dolan(2011)}]{chen2011collecting}
Chen, D.; and Dolan, W.~B. 2011.
\newblock Collecting highly parallel data for paraphrase evaluation.
\newblock In \emph{Proceedings of the 49th annual meeting of the association for computational linguistics: human language technologies}, 190--200.

\bibitem[{Chen et~al.(2025)Chen, Wang, Cao, Liu, Gao, Cui, Zhu, Ye, Tian, Liu, Gu, Wang, Li, Ren, Chen, Luo, Wang, Jiang, Wang, He, Shi, Zhang, Lv, Wang, Shao, Chu, Tu, He, Wu, Deng, Ge, Chen, Zhang, Wang, Dou, Lu, Zhu, Lu, Lin, Qiao, Dai, and Wang}]{chen2025expandingperformanceboundariesopensource}
Chen, Z.; Wang, W.; Cao, Y.; Liu, Y.; Gao, Z.; Cui, E.; Zhu, J.; Ye, S.; Tian, H.; Liu, Z.; Gu, L.; Wang, X.; Li, Q.; Ren, Y.; Chen, Z.; Luo, J.; Wang, J.; Jiang, T.; Wang, B.; He, C.; Shi, B.; Zhang, X.; Lv, H.; Wang, Y.; Shao, W.; Chu, P.; Tu, Z.; He, T.; Wu, Z.; Deng, H.; Ge, J.; Chen, K.; Zhang, K.; Wang, L.; Dou, M.; Lu, L.; Zhu, X.; Lu, T.; Lin, D.; Qiao, Y.; Dai, J.; and Wang, W. 2025.
\newblock Expanding Performance Boundaries of Open-Source Multimodal Models with Model, Data, and Test-Time Scaling.
\newblock arXiv:2412.05271.

\bibitem[{Cheng et~al.(2024)Cheng, Leng, Zhang, Xin, Li, Chen, Zhu, Zhang, Luo, Zhao, and Bing}]{cheng2024videollama2advancingspatialtemporal}
Cheng, Z.; Leng, S.; Zhang, H.; Xin, Y.; Li, X.; Chen, G.; Zhu, Y.; Zhang, W.; Luo, Z.; Zhao, D.; and Bing, L. 2024.
\newblock VideoLLaMA 2: Advancing Spatial-Temporal Modeling and Audio Understanding in Video-LLMs.
\newblock arXiv:2406.07476.

\bibitem[{de~Paula, Salvadeo, and de~Araujo(2022)}]{de2022camnuvem}
de~Paula, D.~D.; Salvadeo, D.~H.; and de~Araujo, D.~M. 2022.
\newblock CamNuvem: A robbery dataset for video anomaly detection.
\newblock \emph{Sensors}, 22(24): 10016.

\bibitem[{Hasan et~al.(2016)Hasan, Choi, Neumann, Roy-Chowdhury, and Davis}]{hasan2016learning}
Hasan, M.; Choi, J.; Neumann, J.; Roy-Chowdhury, A.~K.; and Davis, L.~S. 2016.
\newblock Learning temporal regularity in video sequences.
\newblock In \emph{Proceedings of the IEEE conference on computer vision and pattern recognition}, 733--742.

\bibitem[{Hu et~al.(2021)Hu, Shen, Wallis, Allen-Zhu, Li, Wang, Wang, and Chen}]{hu2021loralowrankadaptationlarge}
Hu, E.~J.; Shen, Y.; Wallis, P.; Allen-Zhu, Z.; Li, Y.; Wang, S.; Wang, L.; and Chen, W. 2021.
\newblock LoRA: Low-Rank Adaptation of Large Language Models.
\newblock arXiv:2106.09685.

\bibitem[{Jin et~al.(2023)Jin, Wang, Alhusaini, Zhao, Liu, Xu, and Zhang}]{jin2023video}
Jin, C.; Wang, T.; Alhusaini, N.; Zhao, S.; Liu, H.; Xu, K.; and Zhang, J. 2023.
\newblock Video fire detection methods based on deep learning: Datasets, methods, and future directions.
\newblock \emph{Fire}, 6(8): 315.

\bibitem[{Kim, Hwang, and Hong(2021)}]{kim2021identifying}
Kim, S.; Hwang, S.; and Hong, S.~H. 2021.
\newblock Identifying shoplifting behaviors and inferring behavior intention based on human action detection and sequence analysis.
\newblock \emph{Advanced Engineering Informatics}, 50: 101399.

\bibitem[{Li et~al.(2022)Li, Li, Xiong, and Hoi}]{li2022blip}
Li, J.; Li, D.; Xiong, C.; and Hoi, S. 2022.
\newblock Blip: Bootstrapping language-image pre-training for unified vision-language understanding and generation.
\newblock In \emph{International conference on machine learning}, 12888--12900. PMLR.

\bibitem[{Lin et~al.(2024)Lin, Ye, Zhu, Cui, Ning, Jin, and Yuan}]{lin2024videollavalearningunitedvisual}
Lin, B.; Ye, Y.; Zhu, B.; Cui, J.; Ning, M.; Jin, P.; and Yuan, L. 2024.
\newblock Video-LLaVA: Learning United Visual Representation by Alignment Before Projection.
\newblock arXiv:2311.10122.

\bibitem[{Liu et~al.(2023)Liu, Li, Wu, and Lee}]{liu2023llava}
Liu, H.; Li, C.; Wu, Q.; and Lee, Y.~J. 2023.
\newblock Visual Instruction Tuning.

\bibitem[{Marr(2010)}]{marr2010vision}
Marr, D. 2010.
\newblock \emph{Vision: A computational investigation into the human representation and processing of visual information}.
\newblock MIT press.

\bibitem[{Mart{\'\i}nez-Mascorro et~al.(2021)Mart{\'\i}nez-Mascorro, Abreu-Pederzini, Ortiz-Bayliss, Garcia-Collantes, and Terashima-Mar{\'\i}n}]{martinez2021criminal}
Mart{\'\i}nez-Mascorro, G.~A.; Abreu-Pederzini, J.~R.; Ortiz-Bayliss, J.~C.; Garcia-Collantes, A.; and Terashima-Mar{\'\i}n, H. 2021.
\newblock Criminal intention detection at early stages of shoplifting cases by using 3D convolutional neural networks.
\newblock \emph{Computation}, 9(2): 24.

\bibitem[{Muneer et~al.(2023)Muneer, Saddique, Habib, and Mohamed}]{muneer2023shoplifting}
Muneer, I.; Saddique, M.; Habib, Z.; and Mohamed, H.~G. 2023.
\newblock Shoplifting detection using hybrid neural network cnn-bilsmt and development of benchmark dataset.
\newblock \emph{Applied Sciences}, 13(14): 8341.

\bibitem[{Nazir et~al.(2023)Nazir, Mitra, Sulieman, and Kamalov}]{nazir2023suspicious}
Nazir, A.; Mitra, R.; Sulieman, H.; and Kamalov, F. 2023.
\newblock Suspicious behavior detection with temporal feature extraction and time-series classification for shoplifting crime prevention.
\newblock \emph{Sensors}, 23(13): 5811.

\bibitem[{Rashvand et~al.(2025{\natexlab{a}})Rashvand, Noghre, Pazho, Ardabili, and Tabkhi}]{rashvand2025shopformer}
Rashvand, N.; Noghre, G.~A.; Pazho, A.~D.; Ardabili, B.~R.; and Tabkhi, H. 2025{\natexlab{a}}.
\newblock Shopformer: Transformer-Based Framework for Detecting Shoplifting via Human Pose.
\newblock In \emph{Proceedings of the Computer Vision and Pattern Recognition Conference}, 5752--5761.

\bibitem[{Rashvand et~al.(2025{\natexlab{b}})Rashvand, Noghre, Pazho, Yao, and Tabkhi}]{rashvand2025exploring}
Rashvand, N.; Noghre, G.~A.; Pazho, A.~D.; Yao, S.; and Tabkhi, H. 2025{\natexlab{b}}.
\newblock Exploring Pose-Based Anomaly Detection for Retail Security: A Real-World Shoplifting Dataset and Benchmark.
\newblock In \emph{Proceedings of the Winter Conference on Applications of Computer Vision}, 1123--1131.

\bibitem[{Reid et~al.(2021)Reid, Coleman, Vance, Kerr, and O’Neill}]{s21206812}
Reid, S.; Coleman, S.; Vance, P.; Kerr, D.; and O’Neill, S. 2021.
\newblock Using Social Signals to Predict Shoplifting: A Transparent Approach to a Sensitive Activity Analysis Problem.
\newblock \emph{Sensors}, 21(20).

\bibitem[{Sultani, Chen, and Shah(2018)}]{sultani2018real}
Sultani, W.; Chen, C.; and Shah, M. 2018.
\newblock Real-world anomaly detection in surveillance videos.
\newblock In \emph{Proceedings of the IEEE conference on computer vision and pattern recognition}, 6479--6488.

\bibitem[{Thakare et~al.(2023)Thakare, Raghuwanshi, Dogra, Choi, and Kim}]{thakare2023dyannet}
Thakare, K.~V.; Raghuwanshi, Y.; Dogra, D.~P.; Choi, H.; and Kim, I.-J. 2023.
\newblock Dyannet: A scene dynamicity guided self-trained video anomaly detection network.
\newblock In \emph{Proceedings of the IEEE/CVF Winter conference on applications of computer vision}, 5541--5550.

\bibitem[{Wan et~al.(2025)Wan, Shen, Liao, Shi, Li, Yang, Zhang, Huang, Zhou, and Yan}]{wan2025qwenlong}
Wan, F.; Shen, W.; Liao, S.; Shi, Y.; Li, C.; Yang, Z.; Zhang, J.; Huang, F.; Zhou, J.; and Yan, M. 2025.
\newblock QwenLong-L1: Towards Long-Context Large Reasoning Models with Reinforcement Learning.
\newblock \emph{arXiv preprint arXiv:2505.17667}.

\bibitem[{Wang and Cherian(2019)}]{wang2019godsgeneralizedoneclassdiscriminative}
Wang, J.; and Cherian, A. 2019.
\newblock GODS: Generalized One-class Discriminative Subspaces for Anomaly Detection.
\newblock arXiv:1908.05884.

\bibitem[{Wu et~al.(2020)Wu, Liu, Shi, Sun, Shao, Wu, and Yang}]{wu2020not}
Wu, P.; Liu, J.; Shi, Y.; Sun, Y.; Shao, F.; Wu, Z.; and Yang, Z. 2020.
\newblock Not only look, but also listen: Learning multimodal violence detection under weak supervision.
\newblock In \emph{European conference on computer vision}, 322--339. Springer.

\bibitem[{Wu et~al.(2024)Wu, Zhou, Pang, Zhou, Yan, Wang, and Zhang}]{wu2024vadclip}
Wu, P.; Zhou, X.; Pang, G.; Zhou, L.; Yan, Q.; Wang, P.; and Zhang, Y. 2024.
\newblock Vadclip: Adapting vision-language models for weakly supervised video anomaly detection.
\newblock In \emph{Proceedings of the AAAI Conference on Artificial Intelligence}, volume~38, 6074--6082.

\bibitem[{Yang et~al.(2025)Yang, Gao, Liu, Wu, Pang, and Shou}]{yang2025assistpda}
Yang, Z.; Gao, C.; Liu, J.; Wu, P.; Pang, G.; and Shou, M.~Z. 2025.
\newblock AssistPDA: An Online Video Surveillance Assistant for Video Anomaly Prediction, Detection, and Analysis.
\newblock \emph{arXiv preprint arXiv:2503.21904}.

\bibitem[{Zanella et~al.(2024)Zanella, Menapace, Mancini, Wang, and Ricci}]{zanella2024harnessing}
Zanella, L.; Menapace, W.; Mancini, M.; Wang, Y.; and Ricci, E. 2024.
\newblock Harnessing Large Language Models for Training-free Video Anomaly Detection.
\newblock In \emph{Proceedings of the IEEE/CVF Conference on Computer Vision and Pattern Recognition}, 18527--18536.

\bibitem[{Zhang et~al.(2024{\natexlab{a}})Zhang, Xu, Wang, Zuo, Han, Huang, Gao, Wang, and Sang}]{zhang2024holmes}
Zhang, H.; Xu, X.; Wang, X.; Zuo, J.; Han, C.; Huang, X.; Gao, C.; Wang, Y.; and Sang, N. 2024{\natexlab{a}}.
\newblock Holmes-vad: Towards unbiased and explainable video anomaly detection via multi-modal llm.
\newblock \emph{arXiv preprint arXiv:2406.12235}.

\bibitem[{Zhang et~al.(2025)Zhang, Xu, Wang, Zuo, Huang, Gao, Zhang, Yu, and Sang}]{zhang2025holmes}
Zhang, H.; Xu, X.; Wang, X.; Zuo, J.; Huang, X.; Gao, C.; Zhang, S.; Yu, L.; and Sang, N. 2025.
\newblock Holmes-vau: Towards long-term video anomaly understanding at any granularity.
\newblock In \emph{Proceedings of the Computer Vision and Pattern Recognition Conference}, 13843--13853.

\bibitem[{Zhang et~al.(2024{\natexlab{b}})Zhang, Li, Liu, Lee, Gui, Fu, Feng, Liu, and Li}]{zhang2024llavanext-video}
Zhang, Y.; Li, B.; Liu, h.; Lee, Y.~j.; Gui, L.; Fu, D.; Feng, J.; Liu, Z.; and Li, C. 2024{\natexlab{b}}.
\newblock LLaVA-NeXT: A Strong Zero-shot Video Understanding Model.

\end{thebibliography}

\clearpage 

\appendix

\section{Appendix A.}
\subsection{Dataset Construction Pipeline.}
\begin{figure}[htb]
\centering
\includegraphics[width=0.9\columnwidth]{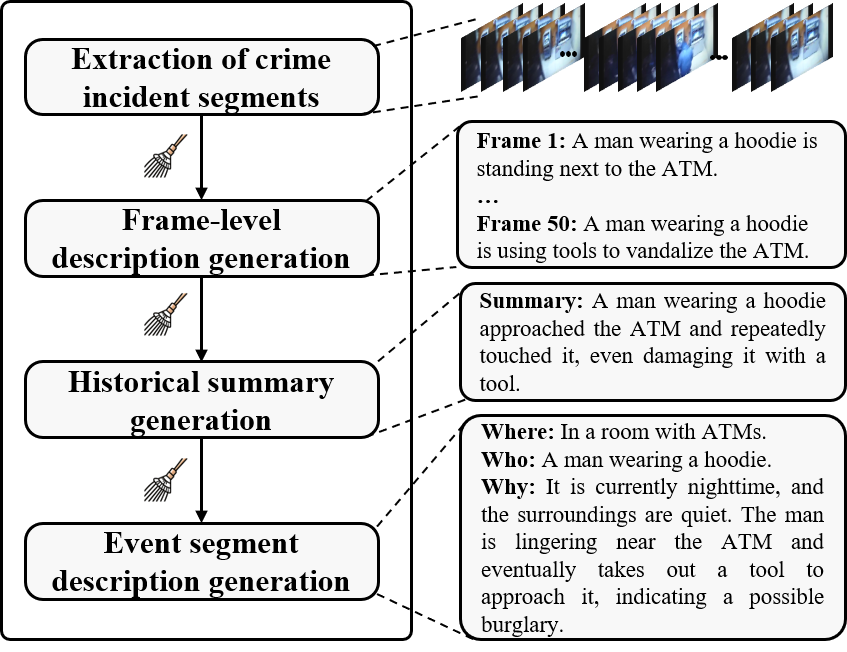}
\caption{Dataset Construction Pipeline.}
\label{figa}
\end{figure}

In this section, we elaborate on the construction pipeline of the dataset, as illustrated in the Figure~\ref{figa}. The pipeline can be divided into three main stages: the frame-level description generation stage, the historical summary generation stage, and the event segment description generation stage.  

\begin{figure}[htb]
\centering
\includegraphics[width=0.95\linewidth]{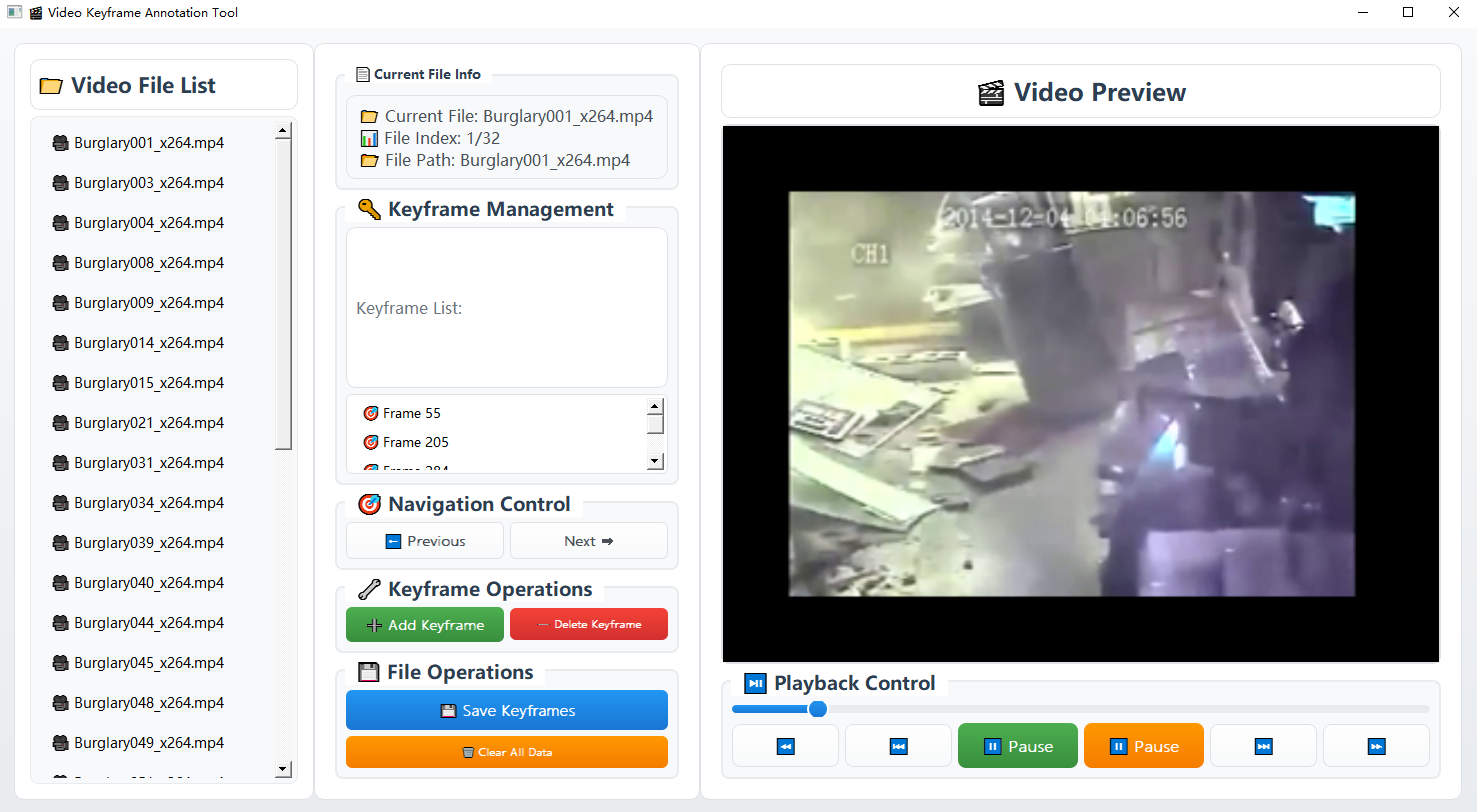}
\caption{Screenshot of the video segments capture interface. Users can browse and load videos through the interface, and annotate key frames and abnormal time periods during real-time playback. All annotation results are dynamically displayed in the interface, providing standardized data support for subsequent model training.}
\label{figb}
\end{figure}

First, in the frame-level description generation stage, video segments containing criminal events with abnormal behaviors are manually extracted, as shown in the Figure~\ref{figb}. From each segment, 100 frames are uniformly sampled and then partitioned into historical and current segments according to different ratios (3:7, 1:1, and 7:3). In this stage, samples with semantically hollow outputs caused by ambiguous image semantics are removed through manual inspection.

Second, in the historical summary generation stage, we employ Qwen-Long to perform semantic integration of the frame-level descriptions from all historical segments, ensuring that each historical segment is associated with a corresponding summary. At this stage, manual calibration is conducted to eliminate hallucinated outputs that result in false subjects or contextually inconsistent behavioral inferences.

Finally, in the event segment description generation stage, GPT-4.1 is utilized to generate textual descriptions that incorporate the location of the event, the subject of the behavior, and the rationale behind the judgment, based on both the historical summary and eight frames from the current segment. This stage also involves manual calibration to ensure that each textual description takes into account both the historical summary information and the visual content of the nearby frames.
\begin{table*}[!ht] 
 \centering 
 \begin{tabular}{lm{14cm}} 
 \hline 
 \textbf{Prompt Type} & \textbf{Prompt Content} \\ 
 \hline 
 Agent 1 & "A picture of [caption]." \\ 
 \hline 
 Agent 2 & "The provided descriptions outline the content of each frame in the video. Your task is to summarize the event in the video based on these descriptions. Identify and analyze whether the same person or object is referred to by different names across frames, such as a 'boy,' 'man,' 'girl,' 'woman,' or 'person' due to possible inconsistencies or ambiguities in the frame descriptions. Do not use bold text, markdown, or bullet points. \textbf{Attention:} (1) Use only factual information provided in the frame descriptions. Do not infer, guess, or supplement missing details; (2) Do not mention or refer to specific frame numbers; (3) Keep the summary concise and clear. If multiple frames convey the same meaning, retain only the description that best expresses the event. You can also use original frame description; (4) The most important: if a weapon such as a gun, knife or bat is mentioned, include only the description of the person holding or using the weapon, or use the original frame description that includes the weapon; (5) The second most important: When multiple frames describe the same person or action, judge whether they refer to the same subject, and merge only if consistent. Do not treat different descriptions blindly as separate people; (6) No line breaks. If many frames describe different things, identify the event that occurs most frequently and is most significant, and summarize that one only; (7) Remove frames with incorrect or inconsistent descriptions based on the context of nearby frames; (8) Ignore any frame with incorrect, unclear, or illogical descriptions. Use contextual judgment to discard them; (9) Write in a coherent, natural sentence. Ensure all parts of the output are logically connected, not disjointed. If the events are too many, output the most important you think; (10) Only describe the important and frequent event. Ignore all others. Be concise and write in a single coherent sentence." \\ 
 \hline 
 Agent 3 & "Focus on the historical texts and image frames. Identify any abnormalities in the following content and provide a reason if any are found." \\ 
 \hline 
 \end{tabular} 
 \caption{Prompt templates used in multi-agent.} 
 \label{taba} 
\end{table*}

\subsection{Multi-Agent Instruction Data Construction.}

In order to construct a dataset for different agents, we designed corresponding prompts based on their specific tasks, including image description, summary, and event description, as shown in Table~\ref{taba}.

\section{Appendix B.}
% \section{Details of Three-Stage Training.}

\subsection{Prompts of Three-Stage Training.}
To effectively guide the Agent 3 in learning criminal behavior prediction, we design a structured three-stage training framework, as shown in the Table~\ref{tabb}. Each stage is driven by a dedicated prompt template, which steers the model to focus on distinct learning objectives.

\begin{table}[!ht]
\centering
\begin{tabular}{lm{5.5cm}} 
\hline
\textbf{Prompt Type} & \textbf{Prompt Content} \\ 
\hline
Stage 1 & 1. "Provide a brief description of the given image." \newline
2. "Write a terse but informative summary of the picture." \newline
3. "Share a concise interpretation of the image provided." \newline
4. "Relay a brief, clear account of the picture shown." \newline
5. "Render a clear and concise summary of the photo." \newline
6. 'Create a compact narrative representing the image presented." \newline
7. "Give a short and clear explanation of the subsequent image." \newline
8. "Create a compact narrative representing the image presented." \newline
9. "Summarize the visual content of the image." \newline
10. "Describe the image concisely." \newline
11. "Offer a succinct explanation of the picture presented." \newline
12. "Present a compact description of the photo's key features." \\
Stage 2 & "Describe the content of the clip in a concise and specific way." \\ 
Stage 3 & "Focus on the historical texts and image frames. Identify any abnormalities in the following content and provide a reason if any are found." \\
\hline
\end{tabular}
\caption{Prompt templates used in three-stage training.}
\label{tabb}
\end{table}

\subsection{Loss Function.}

To train the language generation component, we employ a standard negative log-likelihood loss, denoted as $\mathcal{L}$. This objective function encourages the model to maximize the conditional likelihood of generating the ground-truth token $y_t$ at each time step, given the previously generated tokens $y_{<t}$, the input images $\{I_k\}_{k=1}^8$, and the corresponding query $Q$, Formally, the loss is defined as:

\begin{equation}
\mathcal{L} = -\sum_t \log P\left(y_t \mid y_{<t}, \{I_k\}_{k=1}^8, \underbrace{[In; Id; S]}_{Q}\right),
\label{eqa}
\end{equation}

\noindent where $Q$ represents the concatenation of the instruction $In$, identifier $Id$, and historical summary $S$. This loss function guides the model to learn contextually coherent and semantically relevant sequences by integrating visual and textual cues.

\subsection{Visualization of Training Effects.}

\begin{figure}[htb]
\centering
\includegraphics[width=\linewidth]{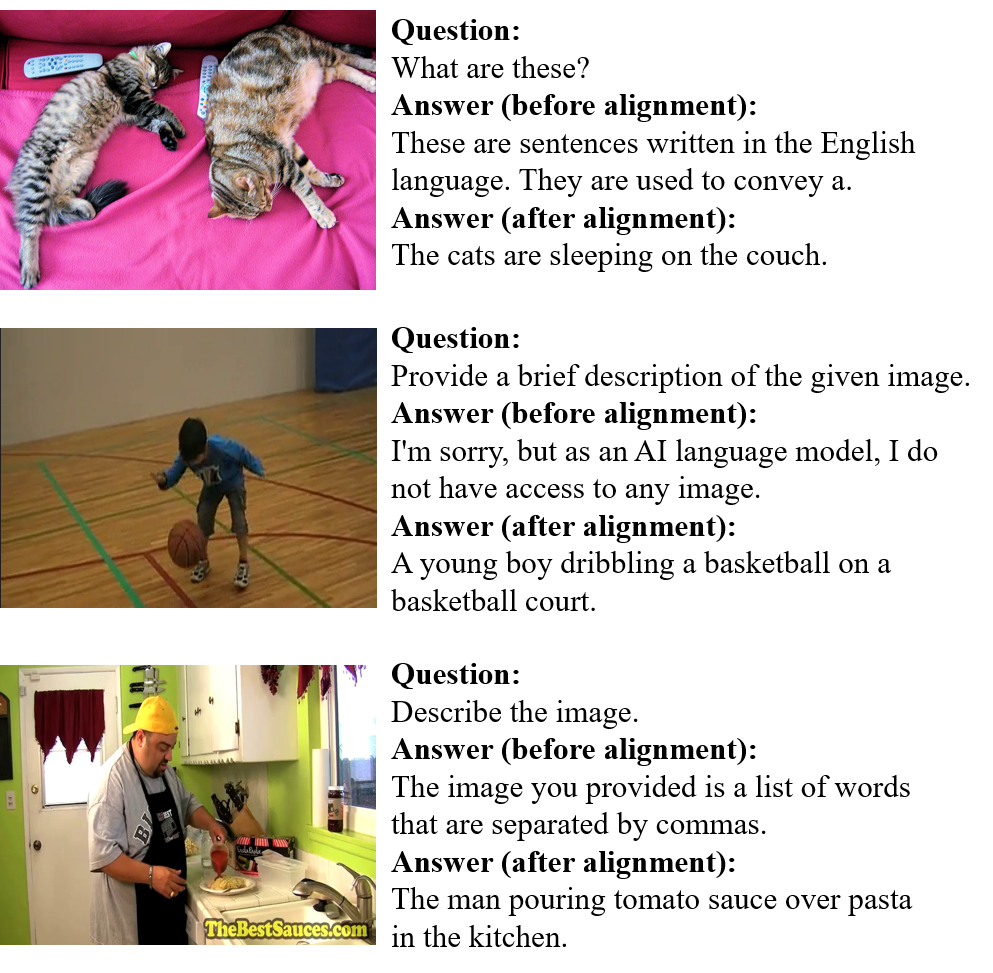}
\caption{Effect of visual-language alignment pre-training.}
\label{figc}
\end{figure}

\begin{figure}[htb]
\centering
\includegraphics[width=\linewidth]{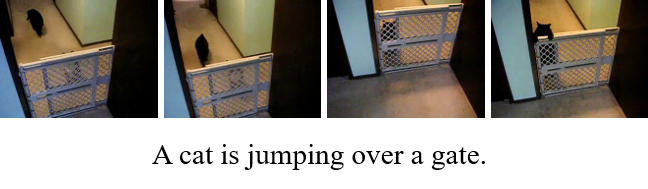}
\caption{Effect of multi-frame semantic modeling fine-tuning.}
\label{figd}
\end{figure}

\begin{figure}[!htb]
\centering
\includegraphics[width=\linewidth]{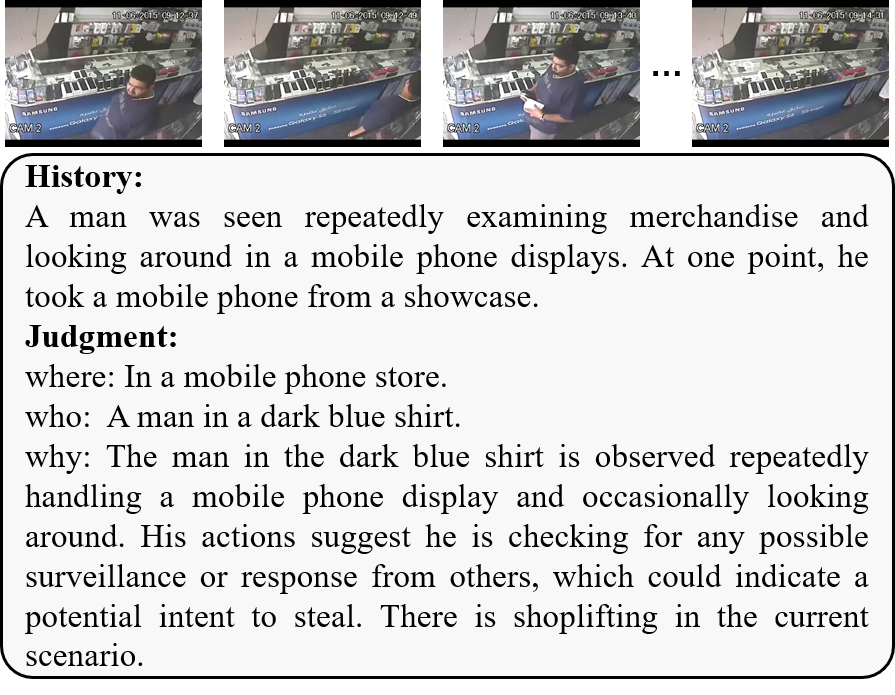}
\caption{Effect of history-enhanced reasoning fine-tuning.}
\label{fige}
\end{figure}

To comprehensively assess the progressive enhancement of the model’s reasoning capability, we qualitatively visualize the outputs obtained after each training stage. The visual-language alignment pre-training stage enables the model to translate visual features into coherent linguistic expressions by aligning image-text pairs, as illustrated in Figure~\ref{figc}. The multi-frame semantic modeling fine-tuning stage captures the evolution of behavioral semantics by modeling consecutive video frames, allowing the model to identify semantic cores, action boundaries, and behavior transitions, as shown in Figure~\ref{figd}. The history-enhanced reasoning fine-tuning stage incorporates the complete temporal context of abnormal video segments to train the model in inferring the underlying intent of actions, as depicted in Figure~\ref{fige}.

\section{Appendix C.}

\subsection{Discussion with Related Works.}

\begin{figure*}[htb]
\centering
\includegraphics[width=0.85\textwidth]{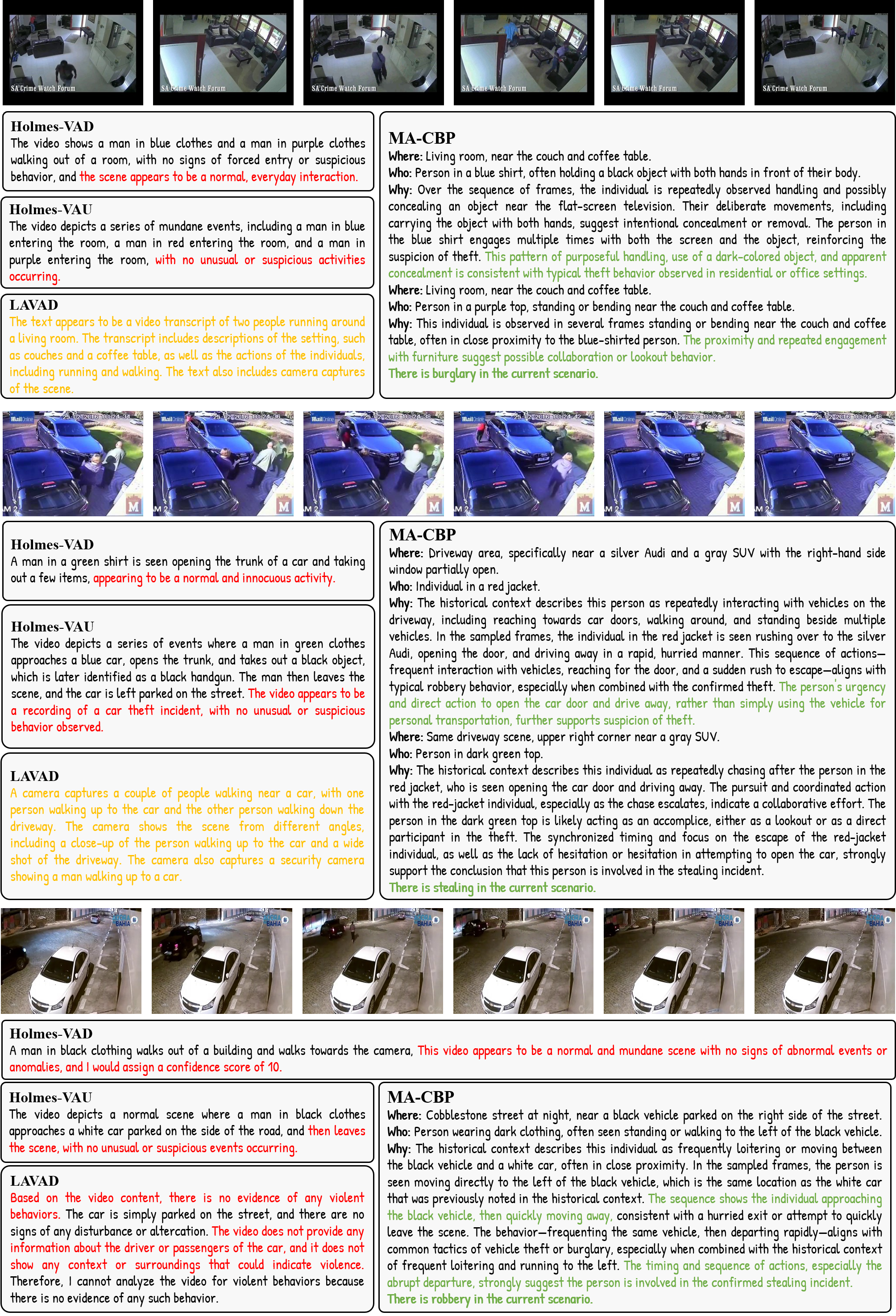}
\caption{Qualitative comparison. Correct, unimportant, and incorrect answers are highlighted in green, yellow, and red, respectively.}
\label{figf}
\end{figure*}

To further demonstrate the advantages of our proposed framework, we conduct a qualitative comparison against several representative video anomaly detection and understanding models, including Holmes-VAD, Holmes-VAU, and LAVAD. As shown in Figure~\ref{figf}. Our model consistently provides more accurate, interpretable, and context-aware descriptions of potentially criminal activities.

Specifically, Holmes-VAD and Holmes-VAU often fail to recognize subtle but significant behavioral cues, such as object concealment, coordinated actions, or abrupt scene departures. Their outputs tend to describe surface-level movements or rely on rule-based assumptions, leading to incorrect judgments. LAVAD, though offering more structured summaries than traditional captioning methods, tends to merely describe what happens without inferring why it happens or whether the observed actions may be suspicious or criminally relevant.

Through multi-agent asynchronous collaboration, we not only describe what happens, but why it happens, by leveraging historical summaries (Agent2) and long-short-term reasoning (Agent3). This enables reasoning over patterns such as deliberate concealment, coordinated actions, or atypical object interactions that LAVAD often ignores.

\section{Appendix D.}
\subsection{Limitations.}

Despite the promising performance of our multi-agent asynchronous framework for criminal behavior prediction, several limitations remain to be addressed. Our current model is constrained by fixed-frame sampling strategies, which may not optimally reflect event granularity or adapt to dynamic video lengths. 

\subsection{Future Work.}
In future work, we plan to explore online inference pipelines that incorporate adaptive frame sampling and temporal attention mechanisms. Moreover, integrating multi-modal cues such as audio or object trajectories may further enhance the robustness of the framework in practical applications.

\end{document}